\documentclass{article}
\usepackage{natbib}
\setcitestyle{round}
\usepackage[affil-it]{authblk}
\usepackage[english]{babel}

\usepackage[english]{babel}

\usepackage[letterpaper,top=2cm,bottom=2cm,left=3cm,right=3cm,marginparwidth=1.75cm]{geometry}

\usepackage{amsmath}
\usepackage{graphicx}
\usepackage{subcaption}
\usepackage{amsfonts}
\usepackage{caption}
\usepackage{float}
\usepackage[colorlinks=true, allcolors=blue]{hyperref}

\title{Exploring Parity Challenges in Reinforcement Learning through Curriculum Learning with Noisy Labels}

\date{}

\author[ ]{Bei Zhou}
\author[ ]{S\o ren Riis\thanks{Corresponding author}}

\affil[ ]{Queen Mary University of London}
\affil[ ]{\textit {\{bei.zhou, s.riis\}@qmul.ac.uk}}

\begin{document}
\maketitle

\begin{abstract}
This paper delves into applying reinforcement learning (RL) in strategy games, particularly those characterized by parity challenges, as seen in specific positions of Go and Chess and a broader range of impartial games. We propose a simulated learning process, structured within a curriculum learning framework and augmented with noisy labels, to mirror the intricacies of self-play learning scenarios. This approach thoroughly analyses how neural networks (NNs) adapt and evolve from elementary to increasingly complex game positions. Our empirical research indicates that even minimal label noise can significantly impede NNs' ability to discern effective strategies, a difficulty that intensifies with the growing complexity of the game positions. These findings underscore the urgent need for advanced methodologies in RL training, specifically tailored to counter the obstacles imposed by noisy evaluations. The development of such methodologies is crucial not only for enhancing NN proficiency in strategy games with significant parity elements but also for broadening the resilience and efficiency of RL systems across diverse and complex environments.

\end{abstract}

\section{Introduction}
\label{sec:parity_introduction}

In combinatorial game theory, impartial games are characterised by positions where players alternate turns, with both players able to move the same set of pieces, thereby having access to the same set of legal moves. The mathematics of this class of games is well-understood. According to the underlying theory for these games, determining the values of the parity function plays a central role in effectively assessing these positions and in identifying advantageous moves.

Nim, a well-known two-player impartial game, exemplifies a key principle in combinatorial game theory. As described by Berlekamp \cite{berlekamp2001winning}, the Sprague–Grundy theorem establishes that any position in an impartial game is mathematically equivalent to a position in Nim. In this game, positions are defined by several heaps, each containing a specific number of counters. Players alternate turns to remove counters from their chosen heaps, aiming to be the last to empty the board and secure a win. This theorem also introduces a method for assigning Grundy numbers, or nimbers, to game positions, which is crucial for evaluating their strategic value. A position is deemed winning if it has a non-zero Grundy number. The essence of Nim’s mathematical complexity lies in determining these nimbers. Specifically, a position’s nimber in Nim is the binary XOR sum of the nimbers of its subpositions, with a zero sum indicating a losing position and a non-zero sum indicating a winning one.

The Nim board positions can be represented by a list of bits, being 1, 0 or -1, where 1 denotes the counters on the board, 0 denotes the counters being removed from the board, and -1 is a token separating the heaps \cite{zhou2022impartial}. To accommodate this Nim board representation, we employ a variant of the parity function, also called parity with noise, which is defined as follows \cite{banino2021pondernet}. Let $n$ be any positive integer. Given input $\{x_1, \ldots ,x_n\}=\{0, 1, -1\}^n$, the function is defined as: 
\begin{equation}
    f({x_1, \ldots ,x_n}) = \left(\sum_{i=1}^n (x_i == 1) \right) \ mod \ 2 
\end{equation} 
where $n$ is the length of the input. The function outputs either 0 or 1, depending on whether the input bitstring contains even or odd numbers of 1s. However, determining if a Nim position is winning is more complicated than figuring out the parity of a board representation. With this or any given board representation, evaluating a Nim board position involves recursively applying the parity function to the number of counters on each heap. Thus, for self-play RL algorithms to master Nim game or other impartial games, they must learn a perfect parity function and know how to apply it to find the winning move. But in some partisan games like Chess or Go, the PV-NNs in AlphaZero do not have to be able to implicitly model a parity function on some positions that involve using the parity function to determine the winning move because after running some MCTS simulations, the best move can still be identified. 

Prior work has concluded that learning a parity function with neural networks from long bitstrings sampled from a uniform distribution is a challenge \cite{cornacchia2023mathematical, abbe2023provable, daniely2020learning, shalev2017failures, raz2018fast}. Recent studies have explored ways to build non-uniform distributions to make it learnable \cite{cornacchia2023mathematical, abbe2023provable, daniely2020learning}. We will summarise their results in Section \ref{sec:parity_related}. 

Noisy labels refer to mislabeled or incorrect labels in the training dataset. In a supervised learning setting, one common approach to addressing noisy labels is to use a surrogate loss function to minimize the impact of these noisy labels and guide the model towards learning the underlying true patterns in the data \cite{song2022learning, scott2015rate, scott2013classification, natarajan2013learning, bylander1994learning}. Typically, these solutions with theoretical assurances require prior knowledge of noise rates. Assuming the label noise is homogeneous, \cite{liu2020peer} developed the peer loss that works without knowing the noise rates. 

In reinforcement learning systems, the reward an agent receives might not reflect its true performance, thus misleading the agent due to the randomness in the reward signal. \cite{wang2020reinforcement} outlined three sources of the noise in the reward function, which are the noise in the environment, task-specific noise and the noise from adversarial attacks. In these cases, measuring the exact noise rates or making assumptions about the noise is extremely difficult. 

\cite{nick2014superintelligence} demonstrated that in reinforcement learning, reward corruption tends to hinder the performance of a wide range of agents and may lead to disastrous consequences for highly intelligent agents.
\cite{everitt2017reinforcement} proved a theorem called \textit{No Free Lunch}, leading to a strong conclusion that no RL agent can outperform a random agent if the noisy rewards are arbitrary and the RL agent does not make any assumption regarding the noise pattern. Building on their results,  \cite{wang2020reinforcement} introduced a reward estimator with which an RL can learn in a noisy environment under perturbed reward, a noisy reward model they came up with. Their RL agent can learn in this scenario where the noises were generated by a reward confusion matrix because the initial condition defined by the \textit{No Free Lunch} theorem is violated. But again, as they pointed out, solving an RL problem where the underlying noise model is unknown and stochastic is a formidable challenge. 

Deep reinforcement learning (DRL) algorithms combine RL algorithms with neural networks that enable them to handle high-dimensional input spaces, exemplified by the AlphaZero algorithm that achieved super-human performances on various board games, including chess, Go and shogi. The AlphaZero algorithm utilises two neural networks called the Policy and Value networks (PV-NNs) \cite{silver2018general}. In the self-play process, the algorithm generates training data by playing against itself. In a self-play game, the states encountered by the agent are labelled as winning states if they won the game; Otherwise, they are treated as losing states \cite{silver2016mastering, silver2017mastering}. In this process, some losing states might be mislabeled as winning if the agent won, and vice versa. This is referred to as the credit assignment problem where the RL agent attributes the reward to the sequence of actions leading to the state where this reward is received. The value network is trained on the actual game output of the self-play simulations \cite{silver2016mastering}, thus making it susceptible to the credit assignment problem. The labels for training the policy network come from the statistics of the Monte-Carlo Tree Search (MCTS) simulations guided by the predictions from the PV-NNs \cite{silver2016mastering}. The incorrect evaluation of the value network could mislead the search and as a consequence, the gathered statistics that the policy network is trained on are also erroneous. 

This credit assignment problem in AlphaZero-style learning, however, is, in general, not a problem as the iterative nature of self-play allows the agent to gradually improve and converge towards more accurate and effective strategies over time. Additionally, it could also bring some extra benefits. The noise in the labels can enhance exploration in the learning process, allowing the algorithm to explore a broader range of moves and preventing it from converging too quickly to a suboptimal solution. However, excessive inaccuracies in the training labels could impede learning progress. 

Our work aims to quantitatively measure to what extent neural networks modelling parity function resist the wrong labels in the dataset across bitstrings of varying length. This could provide more insights into investigating the impact of the credit assignment problem in self-play reinforcement learning algorithms that employ neural networks, like AlphaZero-style algorithms on mastering impartial games. 

Our results show that the appearance of more than 5\% noisy labels impedes the ability of the neural network to model a parity function on long bitstrings (of length 100). This indicates that the wrong labels might challenge self-play RL algorithms to master impartial games with large board positions.  

The structure of this paper is as follows. Section \ref{sec:parity_related} summarises recent studies on the intricacies of modelling a parity function using various neural networks. Section \ref{sec:parity_learning} explains our experiment setup and presents our results on the noisy label problem in a simulated reinforcement learning setting. This paper is concluded in Section \ref{sec:parity_conclusion}. The codes and the data used in this paper are publicly available \footnote{GitHub. \url{https://github.com/sagebei/Exploring-Parity-Challenges-in-Reinforcement-Learning}}. 

\section{Related Work}
\label{sec:parity_related}

Learning the parity function using a neural network is a trivial task on short bitstrings. One of our experiments showed that a single-layer Long Short-Term Memory (LSTM) model with 16 nodes could model a perfect parity function on uniformly sampled bitstrings of length 20. A request of research \footnote{OpenAI. \url{https://openai.com/research/requests-for-research-2}} released by OpenAI proposed a question with regards to using LSTM to model the parity function under two different sets of bitstrings. One database consists of 100,000 random bitstrings of a fixed length of 50, whereas the other one contains the same number of bitstrings, but the length of each bitstring is independently and randomly chosen between 1 and 50. The results were with the first database, the LSTM model failed to learn the parity function, but it succeeded with the second dataset. Additionally, on the bitstrings with varying lengths, the LSTM model tends to converge faster if it was fit in shorter bitstrings than longer ones at the early training stage. This suggests that the performance of the neural network in modelling the parity function is influenced not only by the length of the bitstrings but also by the distribution from which they are sampled.

Recent work has studied the effects of these two factors on neural networks' ability to model a parity function. \cite{shalev2017failures} has proved that under uniformed data, the difficulty grows exponentially as the bitstring length, and this challenge is shared among various neural network architectures. The difficulty, however, can be eased under other specific data distributions or other learning schemes. 

\subsection{Learning parity function with uniform data}

\cite{cornacchia2023mathematical, abbe2023provable, daniely2020learning} have highlighted that modelling a parity function from uniformly sampled long bitstrings is extremely difficult. \cite{shalev2017failures} proved it and showed that this difficulty stems from the insufficient information conveyed by the gradient with respect to the underlying function, which is in this case the parity function. Here we summarise the main results from \cite{shalev2017failures}. 

Let $\mathcal{X} \in \{0, 1\}^n$ be the instance space, and $\mathcal{Y} \in \{0, 1\}$ the label space. Given a dataset $(\boldsymbol{X},\boldsymbol{y})$ where $\mathbf{x} \in \boldsymbol{X}$ is uniformly sampled, let $h \in \mathbf{\mathcal{H}}$ be the underlying function that maps $\boldsymbol{X}$ to $\boldsymbol{y}$, where $\mathbf{\mathcal{H}}$ represents all the $2^n$ possible functions over the instance space. The parity function is one of the functions in $\mathbf{\mathcal{H}}$. We consider a binary classification task that uses a neural network $p_\mathbf{w}$ parameterised by $\mathbf{w}$ to model the target function $h$ from this database via gradient-based methods with loss function $\ell$ being hinge loss or cross-entropy loss. This leads to an optimisation problem that can be formalised as:

\begin{equation}
    \min_{\mathbf{w}}F_h(\textbf{w}) = \min_{\mathbf{w}}\mathbb{E}_{\mathbf{x}}[{\ell}(p_\mathbf{w}(\mathbf{x}), h(\mathbf{x}))]
\end{equation}

The gradient $\nabla F_h(\textbf{w})$ concerning $\textbf{w}$ holds useful information about the target function $h$. They further studied the variance of $\nabla F_h(\textbf{w})$ w.r.t the function $h$ randomly sampled from $\mathbf{\mathcal{H}}$ to measure the expected amount of information about the target function revealed by the gradient, and concluded that: 

\begin{equation}
    \mathrm{Var}(\mathbf{\mathcal{H}}, F, \mathbf{w}) \leq \frac{G(\mathbf{w}^2)}{|\mathbf{\mathcal{H}}|}
    = \frac{C}{2^n}
\end{equation}

where $G(\mathbf{w}^2)$ is constant given $\textbf{w}$. So we denote it as a constant $C$. This variance reduces exponentially as bitstring length increases, indicating that when the length of the bitstring is sufficiently large, the gradient at any point of $\textbf{w}$ will be centred around a fixed point, resulting in trapping the optimisation process in local optima. The variance is also independent of the target function $h$, meaning that this result applies to $h$ being the parity function. They further provided a lower upper bound on the variance, given by
\begin{equation}
    \mathrm{Var}(\mathbf{\mathcal{H}}, F, \mathbf{w}) \leq O(\mathrm{exp}(-\Omega(d)) + \mathrm{exp}(-\Omega(r)))
\end{equation}

where $r$ is norm of $\mathbf{x}$, $||\mathbf{x}||$. This gave rise to the conclusion that any neural network trained under gradient-based methods cannot model the parity function when the bitstring length and $r$ are moderately large.  Their experiments show that the difficulty of learning the parity function grows as the bitstring becomes longer, and their multi-layer perception (MLP) model did not perform any better than random guessing when the length of the bitstring reaches 30.

\subsection{Learning parity function with non-uniform data}
Although learning parity function under uniform data distribution on long bitstrings faces strict constraints, some prior studies have shown that when the bitstrings are not sampled uniformly, neural networks trained by gradient descent methods can model the parity function efficiently with the data sampled from certain distributions. \cite{cornacchia2023mathematical} proposed a new way to build the dataset in which the bitstrings are sampled from two or more product distributions and the neural network is trained using curriculum learning (CL), leading to a significant reduction in the computational cost of learning the parity function. \cite{abbe2023provable} further proved that learning the parity function in the curriculum learning fashion provides certain advantages over the standard training method where data are sampled uniformly. In CL, the model is presented with sparse examples at the early stage of training after which the whole dataset is presented to the model. This enables the model to acquire some knowledge in the beginning and primes it to learn from complex data. 

\cite{daniely2020learning} emphasised learning parities under uniform distribution is computationally challenging and thus constructed and used a dataset sampled from a mixture of two distributions to make the parity function learnable for neural networks using gradient descent methods. 

Pondernet is a new type of neural network tested against modelling the parity function \cite{banino2021pondernet}. To make the parity function learnable, in their setting, the bitstrings of length $n$ are sampled in a way that a random number of bits from 1 to $n$ were randomly set to 1 or -1 and the remaining bits were set to 0. This approach establishes a \textit{latent curriculum} where the model is presented with a mix of sparse and dense bitstrings in each training batch. Here, the density of a bitstring is measured by Hamming weight, where the hardness of a bitstring is defined by the number of 1 and -1 bits in it, analogous to the measure used in  \cite{abbe2023provable}. 

\subsection{Learning parity function with various NN architectures}

Some prior works focused on constructing specific neural networks with fixed weight parameters, mainly MLP or Recurrent Neural Networks (RNN), dedicated to solving the parity problem \cite{hohil1999solving, liu2002n, franco2001generalization, wilamowski2003solving, franco2001generalization}. An RNN with 3 neurons and 12 frozen parameters can approximate an XOR function. These artificially hard-wired neural networks can generalize to all seen and unseen patterns without training or adaptation \cite{al2005neural}. However, it is impossible to artificially set up the weight parameters for neural networks to model the unknown data distributions.

Setting aside the impact of the distribution of bitstrings, most related studies focused on investigating the inherent limitations of various neural networks in modelling the parity function. RNN is an umbrella term that incorporates vanilla (simple) RNN, Bidirectional RNN (BRNN), Gated Recurrent Unit (GRU), LSTM and a wide range of their variations like the Memory-Augmented RNN (MRNN) that enhances RNNs' ability on handling on sequential data with long dependencies \cite{zhao2020rnn}. \cite{zhao2020rnn} argued that RNN and LSTM are incapable of processing long time series data that requires persistent memorization, which is a disadvantage to processing long bitstrings as flipping any single bit alters the parity and a tiny error in the memory might incur disastrous repercussion. This discovery aligns with the theoretical results that the longer bitstrings on which the neural networks are trained, the harder it is for them to learn to model the parity function \cite{hahn2020theoretical}.

The parity of bitstrings is permutation invariant by its nature, as changing the order of the bits does not affect the parity. RNNs, despite being dependent on the order of the input, can be regularized towards permutation invariant. \cite{cohen2020regularizing} showed that RNNs with regularization applied towards permutation invariance can simulate the correct parity function for the bitstrings whose length is up to 100. 

RNN architectures are, in theory, able to simulate any Turing Machine (TM) \cite{al2005neural, siegelmann1995computational} given the suitable weights and biases, but the results are predicated on unrealistic assumptions such as unlimited computation time and infinite precision representation of states. In practice, finding proper parameters through gradient descent algorithms for RNNs is a demanding task due to the notoriously hard {\it vanishing gradient problem} \cite{hochreiter1997long}. With the number of bitstrings increasing, the difficulty of the RNN modelling the parity function also escalates drastically.

Self-Attention networks (a.k.a Transformer architectures) \cite{vaswani2017attention} underpin many Natural Language Processing (NLP) applications since its emergence in 2017. However, \cite{hahn2020theoretical} has shown strong theoretical limitations of the abilities of the Self-Attention network on evaluating logical formulas, which is equivalent to evaluating the parity of bitstrings, and draw a conclusion from asymptotic results that any transformers will make mistakes on modelling the parity when the input is sufficiently long. 

Researchers from DeepMind developed PonderNet \cite{banino2021pondernet} and showed its ability to model the parity function by adaptive computation where the computation needed is proportional to the complexity of the problem. The PonderNet achieved almost perfect accuracy on this hard extrapolation task. Still, unlike RNNs that can process the input vectors of any length through unrolling, it can only be evaluated on the same length of the input vectors as the ones it was trained on. Their experiments were intended to demonstrate the computational adaptability of the PonderNet using the parity function as a testbed rather than exhibiting its power in modelling the parity function. But it shows that as the bitstrings become increasingly complicated, \textit{i.e.} more 1s or -1s in the bitstring \cite{abbe2023provable}, the computational resources required to learn the parity increases to the extent where when the bitstring is sufficiently long, the required resources are enormous, if not unobtainable.

\section{Learning parity function with noisy labels}
\label{sec:parity_learning}

\cite{zhou2022impartial} implemented an AlphaZero-style algorithm for Nim in which any board position is represented by a fixed-length list consisting of 0, 1 and -1s. The initial board position where no counters are removed is represented by a list of 1s and -1s, and the representation of the end position consists of 0s and -1s. During the game, each move results in some 1s being replaced by 0s. Thus the collection of game states encountered by the agent resembles the ones that form a latent curriculum and contains more easy samples compared with the ones generated from the uniform distribution. 

\subsection{Learning parity function from latent curriculum}
Given this observation, our investigation focused on the impact of bitstring length on neural network performance in modelling the parity function within a latent curriculum setting. This setting closely mirrors the optimal scenario in self-play reinforcement learning (RL), where the progression of learning experiences ideally forms an effective curriculum for the agent.
% Given this observation, we investigated how the bitstring length affects the performance of the neural networks in modelling the parity function under the latent curriculum setting which resembles the situation in which a self-play RL agent learns. 
As a single-layer LSTM model can model a perfect parity function, we used it as our model. The model was optimized under binary cross entropy (BCE) loss with mini-batch SGD with size 128. The maximum number of training steps in each experiment is 7.5 million. Our experiments tested bitstrings whose length spans from 20 to 100. We repeated each of the experiments with 10 different random seeds. Under each experiment, the model is considered to have simulated a parity function if its accuracy on test data exceeds 95 per cent. An experiment is conceptually considered successful if the model models the parity function. We will refer to this definition later when presenting the results.

\begin{figure}[H]
\centering
\begin{subfigure}[b]{0.65\textwidth}
\centering
\includegraphics[width=0.65\textwidth]{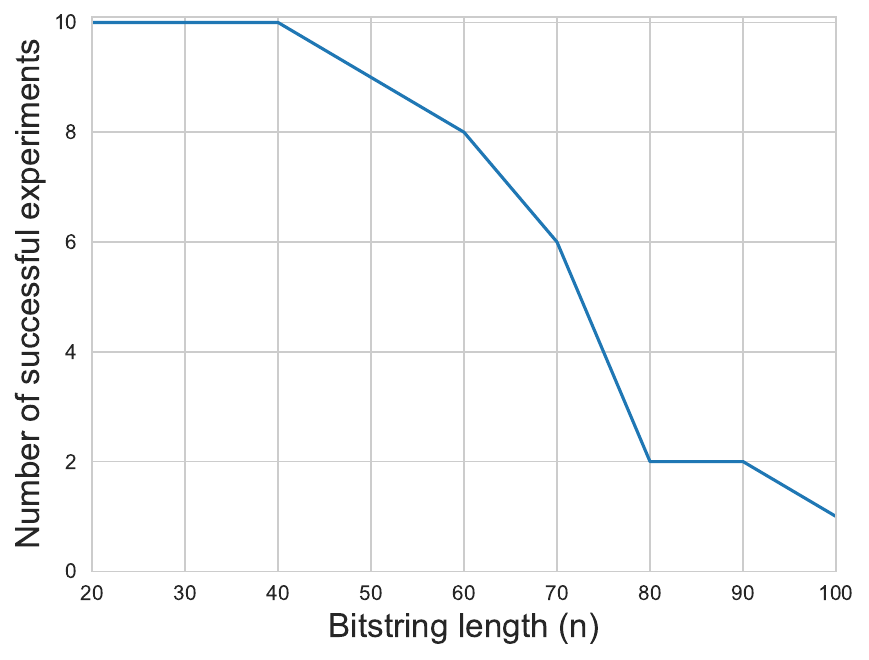}
\caption{The number of experiments in which the LSTM succeeded in modelling the parity function in 7.5 million training steps.}
\label{fig:bitstring_a}
\end{subfigure}

\begin{subfigure}[b]{0.65\textwidth}
\centering
\includegraphics[width=0.65\textwidth]{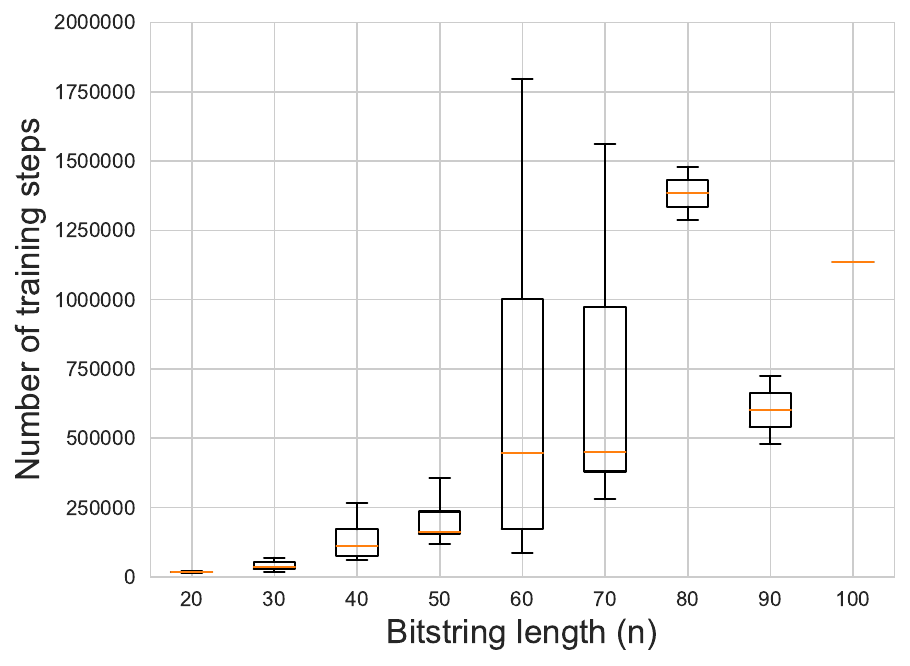}
\caption{The number of training steps needed for the LSTM model to model the parity function from bitstrings of lengths ranging from 20 to 100.}
\label{fig:bitstring_b}
\end{subfigure}

\caption{Under the dataset generated by the latent curriculum, 10 experiments with 10 random seeds were conducted for each given bitstring length.}
\label{fig:bitstring}
\end{figure}

Figure \ref{fig:bitstring} summarises the results of these experiments. Figure \ref{fig:bitstring_a} demonstrates the number of experiments out of 10 in which the model succeeded in modelling a parity function. It is salient that as the length of the bitstring expands, more experiments fail as the longer bitstrings introduce more variance to the latent curriculum. 

Figure \ref{fig:bitstring_b} describes the relationship between the length of bitstrings and the number of training steps the model needed to model the parity function if they succeeded, along with their sensitivity to the nuances of the sampled data under different seeds measured by the difference among the numbers of training steps for a given bitstring length. Considering that in the failed experiments where the number of training steps exceeded 7.5 million, we can conclude that the model is more sensitive to the subtleties of the distribution from which the bitstrings are sampled as their length grows, and additionally, the model tends to need more training steps on longer bitstrings. 

Our results align with the results from \cite{daniely2020learning} that showed on bitstrings sampled from normal distribution, neural networks on longer bitstrings require more training steps to converge. While the training process with the bitstrings constituting a latent curriculum differs from the CL where only sparse samples are presented at the early stage of training as each batch of samples fitting into the NN consists of both sparse and dense samples, our results on the latent curriculum can be extended to a CL setting that mirrors the way the data samples are presented to NNs in a self-play RL algorithm. 

\subsection{Learning parity function from latent curriculum with noisy labels}

We showed in Section \ref{sec:parity_introduction} that in AlphaZero-style training for two-player games, the label of the states arising in the play is determined by the outcome of the games, meaning that the label for the training samples is subject to change and is likely to be inaccurate, especially in the early stage of the training. This counteracts the profits of curriculum learning in which simple samples are presented to facilitate learning in the early training stage. 

We considered an ideal self-play RL setting in which the states an agent acts in establish a latent curriculum, enabling the neural networks in RL to learn a parity function and the noisy label problem is the only problem that could affect the ability of a neural network to model a parity function. To simulate this setting, we propose a new binary classification task in which the data distribution provides a latent curriculum learnable by neural networks via gradient descent methods and the portion of noisy labels is in inverse proportion to the model accuracy. Concretely, the dataset $\mathcal{D}_t$ used at each of the training epochs consists of $\rho$ per cent of data with incorrect noisy labels ($\mathcal{D}_{nl}$), and the rest have correct labels ($\mathcal{D}_{cl}$), as formulated below. 

\begin{equation}
    \mathcal{D}_t = \rho \mathcal{D}_{nl} + (1-\rho) \mathcal{D}_{cl}
\end{equation}

The proportion of noisy labels decreases linearly as the model accuracy increases. Their relationship can be described by the equation below. 

\begin{equation}
    \rho = \begin{cases}
        \rho_0 & x \leq 0.5 \\
        2 \rho_0 (1 - x) & \mathrm{otherwise} \\
    \end{cases}
\end{equation}

where $\rho_0$ represents the percentage of mislabeled data before the training begins, mimicking the noise brought by an untrained model in AlphaZero-style algorithms. This setting enforces that the proportion of noisy data in the dataset decreases linearly from an initial percentage of $\rho_0$ to zero as the model's accuracy on the dataset increases from the initial roughly 50 per cent at the onset of the training to 100 per cent where the neural network models perfect parity function. 

\begin{figure}[H]
    \centering
    \includegraphics[width=0.45\textwidth]{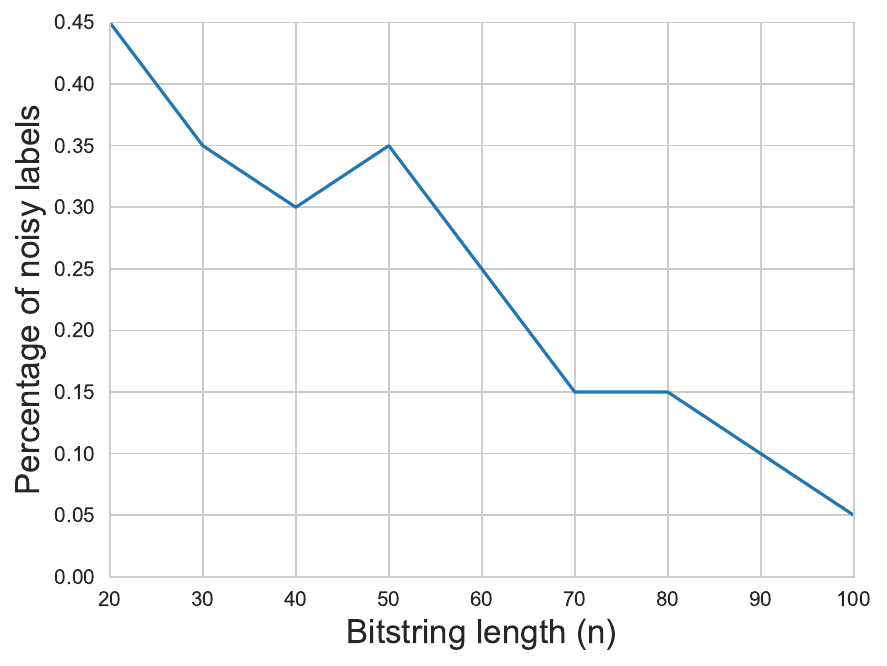}
    \caption{The maximum percentage of the noisy labels in the dataset on which the LSTM model can model the parity function with more than 95\% accuracy across bitstrings whose length ranges from 20 to 100. }
    \label{fig:bitstring_percent_nl}
\end{figure}

We extend our experiments on latent curriculum learning to further test the impact of the noisy labels on the ability of the model to model parity function. As shown in Figure \ref{fig:bitstring_percent_nl}, it is obvious that the model is more susceptible to these mislabels on longer bitstrings. In our setting, the model is immune to mislabels accounting for 45 per cent of the data on the bitstrings of length 20, while more than 5 per cent of wrong labels on bitstrings of length 100 can hinder the capability of the model to simulate a parity function. 

In this simulated ideal self-play RL setting, the noisy labels on long bitstring adversely impact the ability of the neural network to model a parity function. This indicates that in a real self-play RL setting where the states the agent learns from do not establish a hidden curriculum, this problem is exacerbated and leads to more difficulty. This observation is consistent with the findings reported in \cite{zhou2022impartial}, where the performance of a self-play AlphaZero algorithm tailored for Nim notably diminishes as the size of the nim board increases.

% This partially explains the results in \cite{zhou2022impartial} where the performance of a self-play AlphaZero algorithm drastically decreases as the nim board grows larger. 

\section{Conclusion}
\label{sec:parity_conclusion}

In this paper, we first illustrated the essential role of a parity function in determining the winning strategy for impartial games and summarised the results of recent studies on the intricacy of learning a parity function with neural networks. These challenges naturally apply to self-play reinforcement learning algorithms that use neural networks on impartial games. 

We further identified the noisy label problem that could pose another challenge. It would be desirable to investigate how the noisy label problem affects the performance of an agent in an actual self-play RL setting, but due to the multitude of factors impacting its performance, such as the randomness in the RL system and their sensitivity to the choice of hyper-parameters, we instead propose a scenario where except noisy labels all other factors are ruled out. Our main work focused on identifying and analysing the detrimental effect brought by the noisy labels on neural networks learning a parity function from long bitstrings that constitute a latent curriculum, simulating an ideal self-play reinforcement learning setting where noisy labels diminish as the increasing model accuracy. Our findings contribute to the studies investigating the challenges facing self-play RL algorithms on impartial games.  

\section*{Acknowledgment}

This work was funded by the Chinese Scholarship Council (CSC). We appreciate the assistance of the IT Research team at the University of Queen Mary for supporting us in using the Apocrita HPC facility.

\bibliographystyle{plainnat}
\bibliography{sample}

\begin{thebibliography}{31}
\providecommand{\natexlab}[1]{#1}
\providecommand{\url}[1]{\texttt{#1}}
\expandafter\ifx\csname urlstyle\endcsname\relax
  \providecommand{\doi}[1]{doi: #1}\else
  \providecommand{\doi}{doi: \begingroup \urlstyle{rm}\Url}\fi

\bibitem[Abbe et~al.(2023)Abbe, Cornacchia, and Lotfi]{abbe2023provable}
Emmanuel Abbe, Elisabetta Cornacchia, and Aryo Lotfi.
\newblock Provable advantage of curriculum learning on parity targets with mixed inputs.
\newblock \emph{arXiv preprint arXiv:2306.16921}, 2023.

\bibitem[Al-Rawi(2005)]{al2005neural}
Mohammed Al-Rawi.
\newblock A neural network to solve the hybrid n-parity: Learning with generalization issues.
\newblock \emph{Neurocomputing}, 68:\penalty0 273--280, 2005.

\bibitem[Banino et~al.(2021)Banino, Balaguer, and Blundell]{banino2021pondernet}
Andrea Banino, Jan Balaguer, and Charles Blundell.
\newblock Pondernet: Learning to ponder.
\newblock \emph{arXiv preprint arXiv:2107.05407}, 2021.

\bibitem[Berlekamp et~al.(2001)Berlekamp, Conway, and Guy]{berlekamp2001winning}
Elwyn~R Berlekamp, John~H Conway, and Richard~K Guy.
\newblock \emph{Winning ways for your mathematical plays, volume 1}.
\newblock AK Peters/CRC Press, 2001.

\bibitem[Bylander(1994)]{bylander1994learning}
Tom Bylander.
\newblock Learning linear threshold functions in the presence of classification noise.
\newblock In \emph{Proceedings of the seventh annual conference on Computational learning theory}, pages 340--347, 1994.

\bibitem[Cohen-Karlik et~al.(2020)Cohen-Karlik, David, and Globerson]{cohen2020regularizing}
Edo Cohen-Karlik, Avichai~Ben David, and Amir Globerson.
\newblock Regularizing towards permutation invariance in recurrent models.
\newblock \emph{arXiv preprint arXiv:2010.13055}, 2020.

\bibitem[Cornacchia and Mossel(2023)]{cornacchia2023mathematical}
Elisabetta Cornacchia and Elchanan Mossel.
\newblock A mathematical model for curriculum learning for parities.
\newblock 2023.

\bibitem[Daniely and Malach(2020)]{daniely2020learning}
Amit Daniely and Eran Malach.
\newblock Learning parities with neural networks.
\newblock \emph{Advances in Neural Information Processing Systems}, 33, 2020.

\bibitem[Everitt et~al.(2017)Everitt, Krakovna, Orseau, Hutter, and Legg]{everitt2017reinforcement}
Tom Everitt, Victoria Krakovna, Laurent Orseau, Marcus Hutter, and Shane Legg.
\newblock Reinforcement learning with a corrupted reward channel.
\newblock \emph{arXiv preprint arXiv:1705.08417}, 2017.

\bibitem[Franco and Cannas(2001)]{franco2001generalization}
Leonardo Franco and Sergio~A Cannas.
\newblock Generalization properties of modular networks: implementing the parity function.
\newblock \emph{IEEE transactions on neural networks}, 12\penalty0 (6):\penalty0 1306--1313, 2001.

\bibitem[Hahn(2020)]{hahn2020theoretical}
Michael Hahn.
\newblock Theoretical limitations of self-attention in neural sequence models.
\newblock \emph{Transactions of the Association for Computational Linguistics}, 8:\penalty0 156--171, 2020.

\bibitem[Hochreiter and Schmidhuber(1997)]{hochreiter1997long}
Sepp Hochreiter and J{\"u}rgen Schmidhuber.
\newblock Long short-term memory.
\newblock \emph{Neural computation}, 9\penalty0 (8):\penalty0 1735--1780, 1997.

\bibitem[Hohil et~al.(1999)Hohil, Liu, and Smith]{hohil1999solving}
Myron~E Hohil, Derong Liu, and Stanley~H Smith.
\newblock Solving the n-bit parity problem using neural networks.
\newblock \emph{Neural Networks}, 12\penalty0 (9):\penalty0 1321--1323, 1999.

\bibitem[Liu et~al.(2002)Liu, Hohil, and Smith]{liu2002n}
Derong Liu, Myron~E Hohil, and Stanley~H Smith.
\newblock N-bit parity neural networks: new solutions based on linear programming.
\newblock \emph{Neurocomputing}, 48\penalty0 (1-4):\penalty0 477--488, 2002.

\bibitem[Liu and Guo(2020)]{liu2020peer}
Yang Liu and Hongyi Guo.
\newblock Peer loss functions: Learning from noisy labels without knowing noise rates.
\newblock In \emph{International conference on machine learning}, pages 6226--6236. PMLR, 2020.

\bibitem[Natarajan et~al.(2013)Natarajan, Dhillon, Ravikumar, and Tewari]{natarajan2013learning}
Nagarajan Natarajan, Inderjit~S Dhillon, Pradeep~K Ravikumar, and Ambuj Tewari.
\newblock Learning with noisy labels.
\newblock \emph{Advances in neural information processing systems}, 26, 2013.

\bibitem[Nick(2014)]{nick2014superintelligence}
Bostrom Nick.
\newblock Superintelligence: Paths, dangers, strategies.
\newblock 2014.

\bibitem[Raz(2018)]{raz2018fast}
Ran Raz.
\newblock Fast learning requires good memory: A time-space lower bound for parity learning.
\newblock \emph{Journal of the ACM (JACM)}, 66\penalty0 (1):\penalty0 1--18, 2018.

\bibitem[Scott(2015)]{scott2015rate}
Clayton Scott.
\newblock A rate of convergence for mixture proportion estimation, with application to learning from noisy labels.
\newblock In \emph{Artificial Intelligence and Statistics}, pages 838--846. PMLR, 2015.

\bibitem[Scott et~al.(2013)Scott, Blanchard, and Handy]{scott2013classification}
Clayton Scott, Gilles Blanchard, and Gregory Handy.
\newblock Classification with asymmetric label noise: Consistency and maximal denoising.
\newblock In \emph{Conference on learning theory}, pages 489--511. PMLR, 2013.

\bibitem[Shalev-Shwartz et~al.(2017)Shalev-Shwartz, Shamir, and Shammah]{shalev2017failures}
Shai Shalev-Shwartz, Ohad Shamir, and Shaked Shammah.
\newblock Failures of gradient-based deep learning.
\newblock In \emph{International Conference on Machine Learning}, pages 3067--3075. PMLR, 2017.

\bibitem[Siegelmann and Sontag(1995)]{siegelmann1995computational}
Hava~T Siegelmann and Eduardo~D Sontag.
\newblock On the computational power of neural nets.
\newblock \emph{Journal of computer and system sciences}, 50\penalty0 (1):\penalty0 132--150, 1995.

\bibitem[Silver et~al.(2016)Silver, Huang, Maddison, Guez, Sifre, Van Den~Driessche, Schrittwieser, Antonoglou, Panneershelvam, Lanctot, et~al.]{silver2016mastering}
David Silver, Aja Huang, Chris~J Maddison, Arthur Guez, Laurent Sifre, George Van Den~Driessche, Julian Schrittwieser, Ioannis Antonoglou, Veda Panneershelvam, Marc Lanctot, et~al.
\newblock Mastering the game of go with deep neural networks and tree search.
\newblock \emph{nature}, 529\penalty0 (7587):\penalty0 484--489, 2016.

\bibitem[Silver et~al.(2017)Silver, Schrittwieser, Simonyan, Antonoglou, Huang, Guez, Hubert, Baker, Lai, Bolton, et~al.]{silver2017mastering}
David Silver, Julian Schrittwieser, Karen Simonyan, Ioannis Antonoglou, Aja Huang, Arthur Guez, Thomas Hubert, Lucas Baker, Matthew Lai, Adrian Bolton, et~al.
\newblock Mastering the game of go without human knowledge.
\newblock \emph{nature}, 550\penalty0 (7676):\penalty0 354--359, 2017.

\bibitem[Silver et~al.(2018)Silver, Hubert, Schrittwieser, Antonoglou, Lai, Guez, Lanctot, Sifre, Kumaran, Graepel, et~al.]{silver2018general}
David Silver, Thomas Hubert, Julian Schrittwieser, Ioannis Antonoglou, Matthew Lai, Arthur Guez, Marc Lanctot, Laurent Sifre, Dharshan Kumaran, Thore Graepel, et~al.
\newblock A general reinforcement learning algorithm that masters chess, shogi, and go through self-play.
\newblock \emph{Science}, 362\penalty0 (6419):\penalty0 1140--1144, 2018.

\bibitem[Song et~al.(2022)Song, Kim, Park, Shin, and Lee]{song2022learning}
Hwanjun Song, Minseok Kim, Dongmin Park, Yooju Shin, and Jae-Gil Lee.
\newblock Learning from noisy labels with deep neural networks: A survey.
\newblock \emph{IEEE Transactions on Neural Networks and Learning Systems}, 2022.

\bibitem[Vaswani et~al.(2017)Vaswani, Shazeer, Parmar, Uszkoreit, Jones, Gomez, Kaiser, and Polosukhin]{vaswani2017attention}
Ashish Vaswani, Noam Shazeer, Niki Parmar, Jakob Uszkoreit, Llion Jones, Aidan~N Gomez, {\L}ukasz Kaiser, and Illia Polosukhin.
\newblock Attention is all you need.
\newblock In \emph{Advances in neural information processing systems}, pages 5998--6008, 2017.

\bibitem[Wang et~al.(2020)Wang, Liu, and Li]{wang2020reinforcement}
Jingkang Wang, Yang Liu, and Bo~Li.
\newblock Reinforcement learning with perturbed rewards.
\newblock In \emph{Proceedings of the AAAI conference on artificial intelligence}, volume~34, pages 6202--6209, 2020.

\bibitem[Wilamowski et~al.(2003)Wilamowski, Hunter, and Malinowski]{wilamowski2003solving}
Bodgan~M Wilamowski, David Hunter, and Aleksander Malinowski.
\newblock Solving parity-n problems with feedforward neural networks.
\newblock In \emph{Proceedings of the International Joint Conference on Neural Networks, 2003.}, volume~4, pages 2546--2551. IEEE, 2003.

\bibitem[Zhao et~al.(2020)Zhao, Huang, Lv, Duan, Qin, Li, and Tian]{zhao2020rnn}
Jingyu Zhao, Feiqing Huang, Jia Lv, Yanjie Duan, Zhen Qin, Guodong Li, and Guangjian Tian.
\newblock Do rnn and lstm have long memory?
\newblock In \emph{International Conference on Machine Learning}, pages 11365--11375. PMLR, 2020.

\bibitem[Zhou and Riis(2022)]{zhou2022impartial}
Bei Zhou and S{\o}ren Riis.
\newblock Impartial games: A challenge for reinforcement learning.
\newblock \emph{arXiv preprint arXiv:2205.12787}, 2022.

\end{thebibliography}

\end{document}